\documentclass[letter]{article} 
\usepackage{iclr2015,times}
\usepackage{hyperref}
\usepackage{url}
\usepackage{amsmath,amssymb}

\usepackage{subcaption}
\usepackage[utf8]{inputenc}
\usepackage{natbib}

\usepackage[ruled,noend,noline,linesnumbered]{algorithm2e}
\usepackage{graphicx}

\title{A Simple and Efficient Method to Generate Word Sense Representations}

\author{
Luis Nieto Pi\~na\\
Spr\aa kbanken, Department of Swedish\\
University of Gothenburg\\
\texttt{luis.nieto.pina@svenska.gu.se} \\
\And
Richard Johansson\\
Spr\aa kbanken, Department of Swedish\\
University of Gothenburg\\
\texttt{richard.johansson@svenska.gu.se} \\
}

\iclrconference 

\begin{document}

\maketitle

\begin{abstract}

	Distributed representations of words have boosted the performance of many Natural
	Language Processing tasks. However, usually only one representation per word is 
	obtained, not acknowledging the fact that some words have multiple meanings. 
	This has a negative effect on the individual word representations
	and the language model as a whole. In this paper we present a simple model
	that enables recent techniques for building word vectors to represent distinct
	senses of polysemic words. In our assessment of this model 
	we show that it is able to effectively discriminate between words' senses
	and to do so in a computationally efficient manner.

\end{abstract}

\section{Introduction}
\label{sect:intro}

	Distributed representations of words have helped obtain better language 
	models~\citep{bengio2003neural} and improve the performance of many 
	natural language processing applications such as named entity 
	recognition, chunking, paraphrasing, or sentiment 
	classification~\citep{turian2010word,socher2011dynamic,glorot2011domain}.
	Recently, \cite{mikolov2013efficient,mikolov2013distributed} proposed the 
	Skip-gram model, which is able to produce high-quality representations
	from large collections of text in an efficient manner.

	Despite the achievements of distributed representations, polysemy or homonymy are 
	usually disregarded even when word semantics may have a large influence 
	on the models. This results in several distinct senses of one same word 
	sharing a representation, and possibly influencing the representations of 
	words related to those distinct senses under the premise that similar
	words should have similar representations. 

	There have been some recent attempts to address this issue. 
	\cite{reisinger2010multi} propose clustering feature vectors corresponding
	to occurrences of words into a predetermined number of clusters,
	whose centers provide multiple representations for each word,
	in an attempt to capture distinct usages of that word.
	\cite{huang2012improving} cluster context vectors instead, built as 
	weighted average of the words' vectors that surround a certain target word.
	\cite{neelakantan2014efficient} imports the idea of clustering context vectors
	into the Skip-gram model.
	
	We present a simple method for obtaining sense representations
	directly during the Skip-gram training phase. It differs from 
	\cite{neelakantan2014efficient}'s approach in that it does not need to create 
	or maintain clusters to discriminate between senses, leading to a significant
	reduction in the model's complexity. In the following sections we describe
	the model and describe an initial assessment of the results it can produce.
	
\section{Model Description}
\label{sect:model}

\subsection{From Word Forms to Senses}

	The distributed representations for word forms that stem from a Skip-gram~\citep{mikolov2013efficient,mikolov2013distributed} model are built 
	on the premise that, given a certain target word, they should serve to predict its 
	surrounding words in a text. I.e., the training of a Skip-gram model, given a 
	target word $w$,
	is based on maximizing the log-probability of the context words of $w$, $c_1, \dots, c_n$:

		\begin{equation}
		\label{eq:sg_obj}
			\sum_{i = 1}^{n} \log p(c_i | w)
		\end{equation}
		
	The training data usually consists of a large collection of sentences or documents, so that 
	the role of target word $w$ can be iterated over these sequences of words, while the context words $c$
	considered in each case are those that surround $w$ within a window of a certain length, . The objective
	becomes then maximizing the average sum of the log-probabilities from Eq.~\ref{eq:sg_obj}.

	We propose modify this model to include a sense $s$ of the word $w$. Note that
	Eq.~\ref{eq:sg_obj} equals
	
		\begin{equation}
		\label{eq:sg_obj2}
			\log p(c_1, \dots, c_n | w)
		\end{equation}
		
	if we assume the context words $c_i$ to be independent of each other given a target 
	word $w$. The notation in Eq.~\ref{eq:sg_obj2} allows us to consider the Skip-gram 
	as a Na\" ive Bayes	model parameterized by word	embeddings~\citep{mnih2013learning}.
	In this scenario, including a sense would amount then to adding a latent variable $s$, 
	and our model's behaviour given a target word $w$ is to select a sense $s$, 
	which is in its turn used to predict $n$ context words $c_1, \dots, c_n$. Formally:
	
		\begin{equation}
		\label{eq:nb}
			p(s,c_1, \dots, c_n | w) = p(s|w) \cdot p(c_1, \dots, c_n | s) 
								 = p(s|w) \cdot p(c_1|s) \dots p(c_n|s),
		\end{equation}
	
	Thus, our training objective is to maximize the sum of the log-probabilities of 
	context words $c$ given a sense $s$ of the target word $w$ plus the log-probability 
	of the sense $s$ given the target word:
	
		\begin{equation}
		\label{eq:obj}
			\log p(s|w) + \sum_{i = 1}^{n} \log p(c_i | s)
		\end{equation}
		
	We must now consider two distinct vocabularies: $V$ containing all 
	possible word forms (context and target words),	and $S$ containing 
	all possible senses for the words in $V$, with 	sizes $|V|$ and
	$|S|$, resp. Given a pre-set $D \in \mathbb{N}$, our ultimate goal 
	is to obtain $|S|$ dense, real-valued vectors of dimension $D$ that 
	represent the senses in our vocabulary $S$ according to the objective 
	function defined in Eq.~\ref{eq:obj}. 
	
	The neural architecture of the Skip-gram model works with two separate 
	representations for the same vocabulary of words. This double 
	representation is not motivated in the original papers,	but it stems 
	from \verb+word2vec+'s code~\footnote{\url{http://code.google.com/p/word2vec/}} 
	that the model builds separate representations for context and target words, 
	of which the former	constitute the actual output of the system.
	(A note by \cite{goldberg2014word2vec} offers some insight into this subject.)
	We take advantage of this architecture and use one of these two representations 
	to contain senses, rather than word forms: as our model only uses target 
	words $w$ as an intermediate step to select a sense $s$, we only do not need 
	to keep a representation for them. In this way, our model builds a representation 
	of the vocabulary $V$, for the context words, and another for the vocabulary 
	$S$ of senses, which contains the actual output. Note that the representation 
	of context words is only used internally for the purposes of this work, and 
	that context words are word forms; i.e., we only consider senses for the 
	target words.

\subsection{Selecting a Sense}

	In the description of our model above we have considered that for each 
	target word $w$ we are able to select a sense $s$. We now explain the 
	mechanism used for this purpose. The probability of a context word $c_i$
	given a sense $s$, as they appear in the model's objective function
	defined in Eq.~\ref{eq:obj}, $p(c_i|s)$, $\forall i \in [1,n]$, can be 
	calculated using the \textit{softmax} function:
	
		\begin{equation}
		\label{eq:softmax}
			p(c_i|s) = \frac{e^{v_{c_i}^{\intercal} \cdot v_{s}}}
							{\sum_{j=1}^{|V|} e^{v_{c_j}^{\intercal} \cdot v_{s}}}
					 = \frac{e^{v_{c_i}^{\intercal} \cdot v_{s}}}{Z(s)},
		\end{equation}
		
	where $v_{c_i}$ (resp. $v_s$) denotes the vector representing context word $c_i$ 
	(resp. sense $s$), $v^{\intercal}$ denotes the transposed vector $v$, and in the 
	last equality we have used $Z(s)$ to identify the normalizer over all context words.
	With respect to the probability of a sense $s$ given a target word $w$, for simplicity 
	we assume that all senses are equally probable; i.e., $p(s|w) = \frac{1}{K}$ for any 
	of the $K$ senses $s$ of word $w$, $\text{senses}(w)$.

	Using Bayes formula on Eq.~\ref{eq:nb}, we can now obtain the 
	posterior probability of a sense $s$ given the target word $w$ and the context 
	words $c_1, \dots, c_n$:
	
		\begin{equation}
		\label{eq:posterior}
		\begin{aligned}
			p(s|c_1, \dots, c_n, w) & = \frac{p(s|w) \cdot p(c_1, \dots, c_n|s)}
										   {\sum_{s_k \in \text{senses}(w)}p(s_k|w) \cdot p(c_1, \dots, c_n| s_k)} \\
									& = \frac{e^{(c_1 + \cdots + c_n) \cdot s} \cdot Z(s)^{-n}}
										   {\sum_{s_k \in \text{senses}(w)}
										          e^{(c_1 + \cdots + c_n) \cdot s_k} \cdot Z(s_k)^{-n}}
		\end{aligned}
		\end{equation}
	
	During training, thus, given a target word $w$ and context words $c_1, \dots c_n$, 
	the most probable sense $s \in \text{senses}(w)$ is the one that maximizes 
	Eq.~\ref{eq:posterior}. Unfortunately, in most cases it is computationally 
	impractical to explicitly calculate $Z(s)$. From a number of possible approximations, 
	we have empirically found that considering $Z(s)$ to be constant yields 
	the best results; this is not an unreasonable approximation if we expect the context
	word vectors to be densely and evenly spread out in the vector space. 
	Under this assumption, the most probable sense $s$ of $w$ is the one that maximizes
	
		\begin{equation}
		\label{eq:posterior2}
			\frac{e^{(c_1 + \cdots + c_n) \cdot s}}
				 {\sum_{s_k \in \text{senses}(w)} e^{(c_1 + \cdots + c_n) \cdot s_k}}
		\end{equation}
		
	For each word occurrence, we propose to select and train only its most probable sense.
	This approach of \textit{hard sense assignments} is also taken in \cite{neelakantan2014efficient}'s work
	and we follow it here, although it would be interesting to compare it with a \emph{soft}
	updates of all senses of a given word weighted by the probabilities obtained with Eq.~\ref{eq:posterior}.
	
	The training algorithm, thus, iterates over a sequence of words, selecting each one
	in turn as a target word $w$ and its context words as those in a window of a 
	maximum pre-set size. For each target word, a number $K$ of senses $s$ is considered, 
	and the most probable one selected according to Eq.~\ref{eq:posterior2}. (Note that, as the number of senses needs to be informed (using, for example, a lexicon),
	monosemic words need only have one representation.) The selected sense
	$s$ substitutes the target word $w$ in the original Skip-gram model, and any of the
	known techniques used to train it can be subsequently applied to obtain sense representations.
	The training process is drafted in Algorithm~\ref{alg:s2v} using Skip-gram with Negative Sampling.
	
	Negative Sampling \citep{mikolov2013distributed}, based on Noise Contrastive Estimation \citep{mnih2012fast}, is a computationally efficient approximation for the original Skip-gram objective 
	function (Eq.~\ref{eq:sg_obj}). In our implementation it learns the sense representations by 
	sampling $N_{neg}$ words from a noise distribution and using logistic
	regression to distinguish them from a certain context word $c$ of a target word $w$. 
	This process is also illustrated in Algorithm~\ref{alg:s2v}.	
	
	\IncMargin{1em}
	\begin{algorithm}[H]
		\DontPrintSemicolon
 		\KwIn{Sequence of words $w_1, \dots, w_N$, window size $n$, 
 			  learning rate $\alpha$, number of negative words $N_{neg}$}
 		\KwOut{Updated vectors for each sense of words $w_i$, $i = 1, \dots, N$}
 		
 		\For{$t = 1, \dots, N$}{
 			$w = w_i$ \;
 			$K \leftarrow$ number of senses of $w$ \;
 			context($w$) = $\{ c_1, \dots, c_n \, | \, c_i = w_{t + i}, \,\, i = -n, \dots, n, \,\, i \neq 0 \}$ \;
 			\For{$k = 1, \dots, K$}{
 				$p_k = \frac{e^{(v_{c_1} + \cdots + v_{c_n}) \cdot v_{s_k}}}
				 {\sum_{j = 1}^K e^{(v_{c_1} + \cdots + v_{c_n}) \cdot v_{s_k}}}$ \;
 			}
 			$s = \arg\,\max_{k = 1, \dots, K} p_k$ \;
 			\For{$i = 1, \dots, n$}{
 				$f = \frac{1}{1 + e^{v_{c_i} \cdot v_s}}$ \;
 				$g_c = \alpha(1 - f)$ \;
 				Update $v_{c_i}$ with $g_d$ \;
 				$g = g_c$ \;
 				\For{$d = 1, \dots, N_{neg}$}{
 					$c \leftarrow$ word sampled from noise distribution, $c \neq c_i$ \;
 					$f = \frac{1}{1 + e^{v_c \cdot v_s}}$ \;
 					$g_d = -\alpha \cdot f$ \;
 					Update $v_c$ with $g_d$ \;
 					$g = g + g_d$ \;
 				}
 				Update $v_s$ with $g$ \;
 			}
 		}

 	\caption{Selection of senses and training using Skip-gram with Negative Sampling. (Note that $v_x$ denotes the vector representation of word $x$.)}
 	\label{alg:s2v}
	\end{algorithm}
	\DecMargin{1em}

\section{Experiments}
\label{sect:eval}

We trained the model described in Section \ref{sect:model} on Swedish text using a context
window of 10 words and vectors of 200 dimensions.
The model requires the number of senses to be specified for each word;
as a heuristic, we used the number of senses listed in the SALDO
lexicon \citep{borin2013}.

As a training corpus, we created a
corpus of 1 billion words downloaded from Spr{\aa}kbanken, the Swedish language
bank.\footnote{\url{http://spraakbanken.gu.se}} The corpora are
distributed in a format where the text has been tokenized,
part-of-speech-tagged and
lemmatized. Compounds have been segmented automatically and when a
lemma was not listed in SALDO, we used the parts of the compounds
instead. The input to the software computing the embeddings
consisted of lemma forms with concatenated part-of-speech tags,
e.g. \emph{dricka}-verb for the verb `to drink' and \emph{dricka}-noun
for the noun `drink'.

The training time of our model on this corpus was 22 hours. For the sake of
time performance comparison, we run an off-the-shelf \verb+word2vec+ 
execution on our corpus using the same parameterization described above; 
the training of word vectors took 20 hours, which illustrates
the little complexity that our model adds to the original Skip-gram.

\subsection{Inspection of nearest neighbors}

We evaluate the output of the algorithm qualitatively by inspecting
the nearest neighbors of the senses of a number of example 
words, and comparing them to the senses listed in SALDO. We leave a
quantitative evaluation to future work.

Table \ref{table:nns} shows the nearest neighbor lists of the senses
of two words where the algorithm has been 
able to learn the distinctions used in the lexicon.
The verb \emph{flyga} `to fly' has two senses listed in SALDO: to travel
by airplane and to move through the air.
The adjective \emph{öm} `tender' also has two senses, similar to the
corresponding English word: one emotional and one physical.
The lists are semantically coherent, although we note that they are
topical rather than substitutional; this is expected since the algorithm was
applied to lemmatized and compound-segmented text and we use a fairly
wide context window.

\begin{table}[htbp]
\centering
\begin{subtable}{.5\textwidth}
\centering
\footnotesize
\begin{tabular}{ll}
\emph{flyg} `flight' & \emph{flaxa} `to flap wings' \\
\emph{flygning} `flight' & \emph{studsa} `to bounce' \\
\emph{flygplan} `airplane' & \emph{sväva} `to hover'\\
\emph{charterplan} `charter plane' & \emph{skjuta} `to shoot'\\
\emph{SAS-plan} `SAS plane' & \emph{susa} `to whiz'
\end{tabular}
\caption{\emph{flyga} `to fly'}
\end{subtable}%
\begin{subtable}{.5\textwidth}
\centering
\footnotesize
\begin{tabular}{ll}
\emph{kärleksfull} `loving' & \emph{svullen} `swollen' \\
\emph{ömsint} `tender' & \emph{ömma} `to be sore' \\
\emph{smek} `caress' & \emph{värka} `to ache'\\
\emph{kärleksord} `word of love' & \emph{mörbulta} `to bruise'\\
\emph{ömtålig} `delicate' & \emph{ont} `pain'
\end{tabular}
\caption{\emph{öm} `tender'}
\end{subtable}

\caption{Examples of nearest neighbors of the two senses of two
  example words.}
\label{table:nns}
\end{table}

\begin{figure}
\centering
\fbox{\includegraphics[scale=0.4]{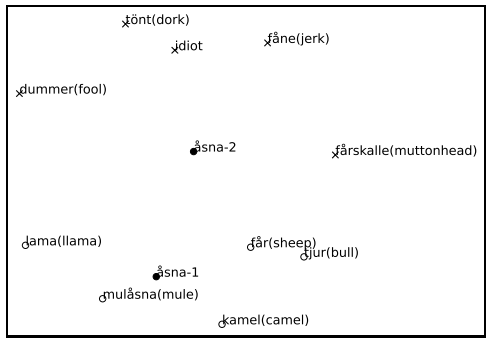}}
\caption{2D projections of the two senses of \emph{åsna} ('donkey' and 'slow-witted person') and their nearest neighbors.}
\label{fig:asna}
\end{figure}

In a related example, Figure~\ref{fig:asna} shows the projections onto a 2D space~\footnote{The projection was computed using \texttt{scikit-learn} \citep{pedregosa2011} using multidimensional scaling of the distances in a 200-dimensional vector space.} of the representations for the two senses of \emph{åsna}: 'donkey' or 'slow-witted person', and those of their corresponding nearest neighbors.

For some other words we have inspected, we fail to find one or more of
the senses. This is typically when one sense is very dominant,
drowning out the rare senses. For instance, the word \emph{rock} has
two senses, `rock music' and `coat', where the first one is much more
frequent. While one of the induced
senses is close to some pieces of clothing, most of its nearest
neighbors are styles of music.

In other cases, the algorithm has come up with meaningful sense
distinctions, but not exactly as in the lexicon.
For instance, the lexicon lists two senses for the noun \emph{böna}:
`bean' and `girl'; the algorithm has instead created two bean senses:
bean as a plant part or bean as food.
In some other cases, the algorithm finds genre-related distinctions instead
of sense distinctions. For instance, for the verb \emph{älska},
with two senses `to love' or `to make love', the algorithm has found
two stylistically different uses of the 
first sense: one standard, and one related to informal words
frequently used in social media. Similarly, for the noun \emph{svamp}
`sponge' or `mushroom'/`fungus', the algorithm does not find the
sponge sense but distinguishes taxonomic, cooking-related, and
nature-related uses of the mushroom/fungus sense.
It's also worth mentioning that when some frequent foreign word is
homographic with a Swedish word, it tends to be assigned to a sense.
For instance, for the adjective \emph{sur} `sour', the
lexicon lists one taste and one chemical sense; the algorithm
conflates those two senses but creates a sense for the French
preposition.

\section{Conclusions and Future Work}

In this paper, we present a model for automatically building sense 
vectors based on the Skip-gram method. 
In order to learn the sense vectors, we modify the Skip-gram model
to take into account the number of senses of each target word. By including 
a mechanism to select the most probable sense given a target word 
and its context, only slight modifications to the original training 
algorithm are necessary for it to learn distinct representations of 
word senses from unstructured text.

To evaluate our model we train it on a 1-billion-word Swedish corpus
and use the SALDO lexicon to inform the number of senses associated to
each word. Over a series of examples in which we analyse the nearest neighbors
of some of the represented senses, we show how the obtained sense 
representations are able to replicate the senses defined in SALDO, or to make
novel sense distinctions in others. On instances in which a sense is 
dominant we observe that the obtained representations favour this sense
in detriment of other, less common ones.

We have used a lexicon just for setting the number of senses of a given word, 
and showed that with that information we are able to obtain coherent sense 
representations. An interesting line of research lies in further 
exploiting existing knowledge resources for learning better sense vectors. 
E.g., integrating in this model the network topology of a lexicon such as 
SALDO, that links together senses of 
related words, could arguably help improve the representations for those 
rare senses with which our model currently struggles by learning their
representations taking into account those of neighboring senses.

We also hope to provide a more systematic evaluation of 
our model so that a more accurate assessment of its qualities can 
be made, and its performance more easily compared against 
that of similar recent work.

\bibliography{representations}

\begin{thebibliography}{14}
\providecommand{\natexlab}[1]{#1}
\providecommand{\url}[1]{\texttt{#1}}
\expandafter\ifx\csname urlstyle\endcsname\relax
  \providecommand{\doi}[1]{doi: #1}\else
  \providecommand{\doi}{doi: \begingroup \urlstyle{rm}\Url}\fi

\bibitem[Bengio et~al.(2003)Bengio, Ducharme, Vincent, and
  Jauvin]{bengio2003neural}
Bengio, Yoshua, Ducharme, R{\'e}jean, Vincent, Pascal, and Jauvin, Christian.
\newblock A neural probabilistic language model.
\newblock \emph{Journal of Machine Learning Research}, 3:\penalty0 1137--1155,
  2003.

\bibitem[Borin et~al.(2013)Borin, Forsberg, and L\"onngren]{borin2013}
Borin, Lars, Forsberg, Markus, and L\"onngren, Lennart.
\newblock {SALDO}: a touch of yin to {WordNet}'s yang.
\newblock \emph{Language Resources and Evaluation}, 47:\penalty0 1191--1211,
  2013.

\bibitem[Glorot et~al.(2011)Glorot, Bordes, and Bengio]{glorot2011domain}
Glorot, Xavier, Bordes, Antoine, and Bengio, Yoshua.
\newblock Domain adaptation for large-scale sentiment classification: A deep
  learning approach.
\newblock In \emph{Proceedings of the 28th International Conference on Machine
  Learning (ICML-11)}, pp.\  513--520, 2011.

\bibitem[Goldberg \& Levy(2014)Goldberg and Levy]{goldberg2014word2vec}
Goldberg, Yoav and Levy, Omer.
\newblock word2vec explained: deriving {Mikolov} et al.'s negative-sampling
  word-embedding method.
\newblock \emph{arXiv preprint arXiv:1402.3722}, 2014.

\bibitem[Huang et~al.(2012)Huang, Socher, Manning, and Ng]{huang2012improving}
Huang, Eric~H, Socher, Richard, Manning, Christopher~D, and Ng, Andrew~Y.
\newblock Improving word representations via global context and multiple word
  prototypes.
\newblock In \emph{Proceedings of the 50th Annual Meeting of the Association
  for Computational Linguistics: Long Papers-Volume 1}, pp.\  873--882.
  Association for Computational Linguistics, 2012.

\bibitem[Mikolov et~al.(2013{\natexlab{a}})Mikolov, Chen, Corrado, and
  Dean]{mikolov2013efficient}
Mikolov, Tomas, Chen, Kai, Corrado, Greg, and Dean, Jeffrey.
\newblock Efficient estimation of word representations in vector space.
\newblock \emph{arXiv preprint arXiv:1301.3781}, 2013{\natexlab{a}}.

\bibitem[Mikolov et~al.(2013{\natexlab{b}})Mikolov, Sutskever, Chen, Corrado,
  and Dean]{mikolov2013distributed}
Mikolov, Tomas, Sutskever, Ilya, Chen, Kai, Corrado, Greg~S, and Dean, Jeff.
\newblock Distributed representations of words and phrases and their
  compositionality.
\newblock In \emph{Advances in Neural Information Processing Systems}, pp.\
  3111--3119, 2013{\natexlab{b}}.

\bibitem[Mnih \& Kavukcuoglu(2013)Mnih and Kavukcuoglu]{mnih2013learning}
Mnih, Andriy and Kavukcuoglu, Koray.
\newblock Learning word embeddings efficiently with noise-contrastive
  estimation.
\newblock In \emph{Advances in Neural Information Processing Systems}, pp.\
  2265--2273, 2013.

\bibitem[Mnih \& Teh(2012)Mnih and Teh]{mnih2012fast}
Mnih, Andriy and Teh, Yee~Whye.
\newblock A fast and simple algorithm for training neural probabilistic
  language models.
\newblock \emph{arXiv preprint arXiv:1206.6426}, 2012.

\bibitem[Neelakantan et~al.(2014)Neelakantan, Shankar, Passos, and
  McCallum]{neelakantan2014efficient}
Neelakantan, Arvind, Shankar, Jeevan, Passos, Alexandre, and McCallum, Andrew.
\newblock Efficient non-parametric estimation of multiple embeddings per word
  in vector space.
\newblock In \emph{Proceedings of EMNLP}, 2014.

\bibitem[Pedregosa et~al.(2011)Pedregosa, Varoquaux, Gramfort, Michel, Thirion,
  Grisel, Blondel, Prettenhofer, Weiss, Dubourg, VanderPlas, Passos,
  Cournapeau, Brucher, Perrot, and Duchesnay]{pedregosa2011}
Pedregosa, Fabian, Varoquaux, Ga{\"e}l, Gramfort, Alexandre, Michel, Vincent,
  Thirion, Bertrand, Grisel, Olivier, Blondel, Mathieu, Prettenhofer, Peter,
  Weiss, Ron, Dubourg, Vincent, VanderPlas, Jake, Passos, Alexandre,
  Cournapeau, David, Brucher, Matthieu, Perrot, Matthieu, and Duchesnay,
  Edouard.
\newblock Scikit-learn: Machine learning in {P}ython.
\newblock \emph{Journal of Machine Learning Research}, 12:\penalty0 2825--2830,
  2011.

\bibitem[Reisinger \& Mooney(2010)Reisinger and Mooney]{reisinger2010multi}
Reisinger, Joseph and Mooney, Raymond~J.
\newblock Multi-prototype vector-space models of word meaning.
\newblock In \emph{Human Language Technologies: The 2010 Annual Conference of
  the North American Chapter of the Association for Computational Linguistics},
  pp.\  109--117. Association for Computational Linguistics, 2010.

\bibitem[Socher et~al.(2011)Socher, Huang, Pennin, Manning, and
  Ng]{socher2011dynamic}
Socher, Richard, Huang, Eric~H, Pennin, Jeffrey, Manning, Christopher~D, and
  Ng, Andrew~Y.
\newblock Dynamic pooling and unfolding recursive autoencoders for paraphrase
  detection.
\newblock In \emph{Advances in Neural Information Processing Systems}, pp.\
  801--809, 2011.

\bibitem[Turian et~al.(2010)Turian, Ratinov, and Bengio]{turian2010word}
Turian, Joseph, Ratinov, Lev, and Bengio, Yoshua.
\newblock Word representations: a simple and general method for semi-supervised
  learning.
\newblock In \emph{Proceedings of the 48th Annual Meeting of the Association
  for Computational Linguistics}, pp.\  384--394. Association for Computational
  Linguistics, 2010.

\end{thebibliography}
\bibliographystyle{iclr2015}

\end{document}